\documentclass[10pt,twocolumn,letterpaper]{article}

\usepackage{iccv}
\usepackage{times}
\usepackage{epsfig}
\usepackage{graphicx}
\usepackage{amsmath}
\usepackage{amssymb}

\graphicspath{{./fig/}}
\DeclareGraphicsExtensions{.pdf}

\usepackage{tabularx}
\usepackage{tabu}
\usepackage{booktabs}
\usepackage{siunitx}
\usepackage{multirow}
\usepackage{float}

\newcolumntype{Y}{>{\raggedleft\arraybackslash}X}
\newcommand{\xw}[1]{\textcolor{blue}{#1}}
\newcommand{\xwc}[1]{\textcolor{red}{\bf [Comments: #1] }}

\usepackage[pagebackref=true,breaklinks=true,letterpaper=true,colorlinks,bookmarks=false]{hyperref}

\iccvfinalcopy % *** Uncomment this line for the final submission

\def\iccvPaperID{3185} % *** Enter the ICCV Paper ID here

\ificcvfinal\fi
\begin{document}

%%%%%%%%% TITLE
\title{Not All Parts Are Created Equal: \\
3D Pose Estimation by Modeling Bi-directional Dependencies of Body Parts}

\author{Jue Wang$^1$\\
% University of Technology Sydney\\
% {\tt\small jue.wang.0911@gmail.com}
\and
Shaoli Wang$^2$\\
% University of Sydney\\
% {\tt\small shaoli.huang@sydney.edu.au}
\and
Xinchao Wang$^3$\\
% Stevens Institute of Technology\\
% {\tt\small xinchao.w@gmail.com}
\and
Dacheng Tao$^2$\\
\and
$^1$University of Technology Sydney\\
$^2$University of Sydney\\
$^3$Stevens Institute of Technology\\
}

\maketitle
%\thispagestyle{empty}

%%%%%%%%% ABSTRACT
\begin{abstract}
    Not all the human body parts have the same~degree of freedom~(DOF) due to the  physiological structure. For example, the limbs may move more flexibly and freely than the torso does. Most of the existing 3D pose estimation methods, despite the very promising results achieved, treat the body joints equally and consequently often lead to larger reconstruction errors on the  limbs. In this paper, we propose a progressive approach that explicitly accounts for the distinct DOFs among the body parts. We model parts with higher DOFs like the elbows, as  dependent components of the corresponding parts with lower DOFs like the torso, of which the 3D locations can be more reliably estimated. Meanwhile, the high-DOF parts may in turn impose a constraint on where the low-DOF ones lie. As a result, parts with different DOFs supervise one another, yielding physically constrained and plausible pose-estimation results. To further facilitate the prediction of the high-DOF parts, we introduce a pose-attribute estimation, where the relative location of a limb joint with respect to the torso, which has the least DOF of a human body, is explicitly estimated and further fed to the joint-estimation module. 
    The proposed approach achieves very promising results, outperforming the state of the art on several benchmarks.

\end{abstract}

%%%%%%%%% BODY TEXT
\section{Introduction}
The unique physiological structure of a human body
results in that different body parts may have different
degrees of freedom~(DOFs).
For example, the motion range 
of a human wrist is significantly broader than that
of a shoulder. Such distinct DOFs further lead to
the varying levels of difficulties when it comes
to 3D pose estimation, for which the goal
is to  predict the 3D locations 
of human body joints from one or multiple images.
%It is a intrinsically demanding problem, due to the 
%ambiguity of projecting 3D human bodies to 2D that leads
%to challenges such as self-occlusions.

    \begin{figure}[t]
        \centering
        \includegraphics[width=0.45\textwidth]{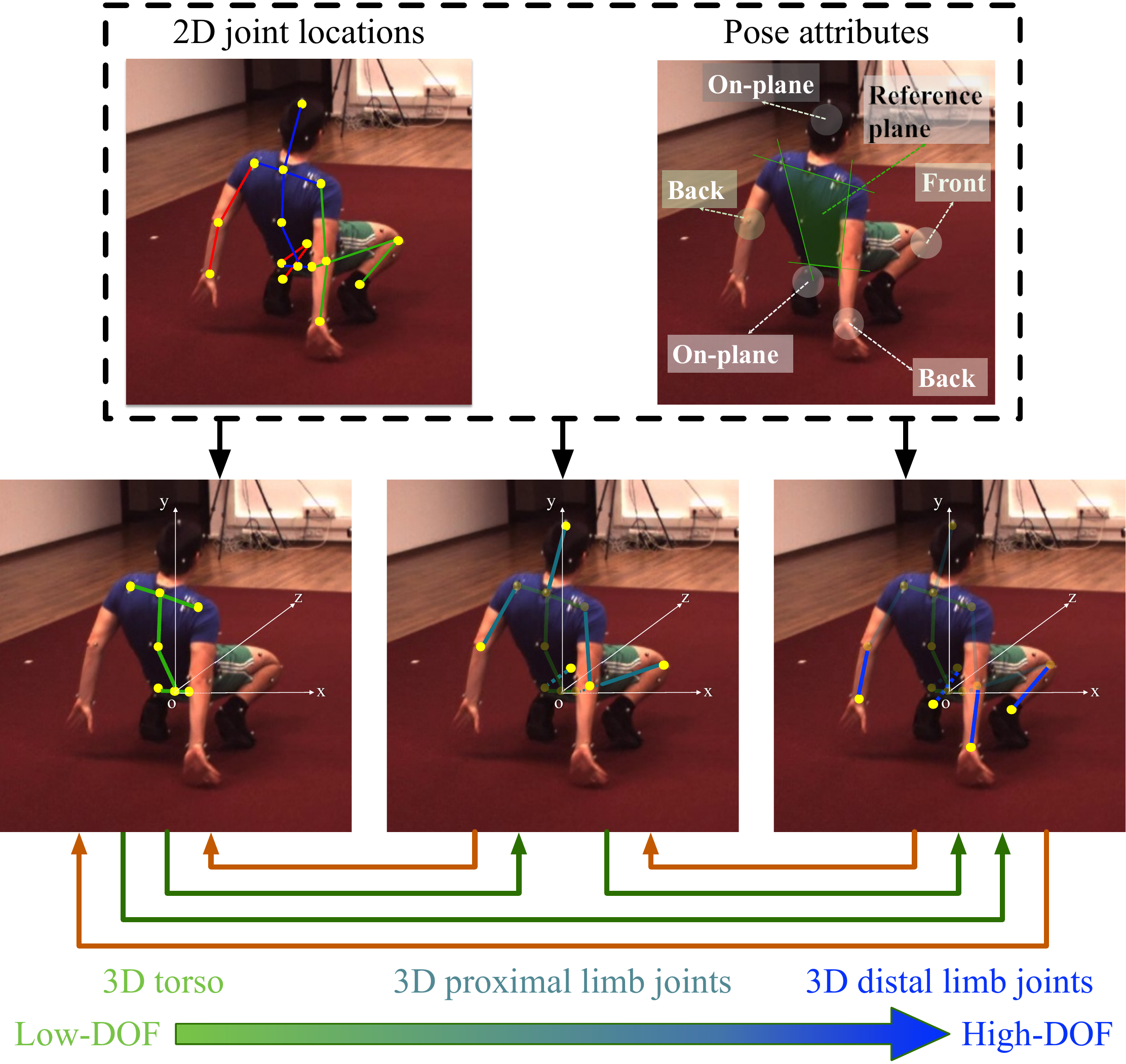}
        \caption{
        Illustration of the proposed approach.
        The 2D joint locations along with the pose attributes
        are first estimated using a multi-task network,
        and   then fed as input to the 
        3D pose estimation network that explicitly
        models the bi-directional dependencies
        among body parts of different DOFs.
        Specifically, the high-DOF body parts are treated as dependent components of low-DOF ones,
        and in turn, provide a constraint on where the low-DOF ones lie.}
        \label{fig_title}
        \vspace{-5mm}
    \end{figure} 

Most of the  existing 3D pose estimation methods~\cite{zhou2016sparseness,zhou2017sparse,8316924,martinez2017simple,moreno20173d,fang2018learning,pavlakos2017coarse,tekin2017learning,zhou2017towards,Pavlakos_2018_CVPR,Dabral_2018_ECCV,Sun_2018_ECCV},
despite the
impressive state-of-the-art performances
achieved,
have overlooked such DOF distinctions among the body parts
and have treated them equally during the learning process.
This consequently leads to the often larger reconstruction errors 
on the more flexible and thus more 
challenging body parts, such as the limbs. 

In this paper, we propose a dedicated 
approach that explicitly
utilize the DOF differences among body parts to 
facilitate 3D pose estimation~(see Fig.~\ref{fig_title}). 
We categorize the body parts into three groups
based on the increasing levels of DOFs: the \emph{torso}, 
the \emph{proximal limb joints} including 
the head, the elbows and the knees,
and the \emph{distal limb joints} 
including the wrists and ankles. 
By such categorization, 
we explicitly model the location 
of a higher-DOF joint like an elbow,
as a dependent variable of a lower-DOF one like the torso
in a progressive manner, 
where the latter can be in most cases estimated
more reliably. 
In turn, the locations of the higher-DOF joints may 
constrain those of lower-DOF ones and potentially
revise the prediction errors. 
Such bi-directional and progressive dependencies enable that
body parts of various DOFs supervise 
one another, yielding 
physically plausible 3D pose-estimation results.

To  further take  advantage of image 
evidences for the dependency modeling, 
we introduce  \emph{pose attribute}, 
which captures the relative location of a limb 
joint with respect to the torso, the 
body part with the least DOF.
Each limb joint may be assigned one of three 
pose attributes, \emph{front}, \emph{back}, and \emph{on-plane},
describing its offset to the torso. 
Given an input image, 
these attributes are estimated together with the
2D pose via a multi-task network, 
and  are further fed to the succeeding 3D pose module that 
explicitly accounts for the aforementioned progressive 
dependencies. In other words, the estimated pose attributes,
being an input to 3D pose estimation, 
provide an explicit and strong prior of where the limb 
may lie, benefiting the downstream part-dependency modeling.
Unlike the regression-based depth estimation that often yields
deviations, which may further propagate to the 3D pose estimation
and deteriorate the results,
the three-class pose-attribute prediction,
as demonstrated in our experiments, turns out to be in most cases
reliable,  providing advantageous image cues.

Our main contributions are thus summarized as follows. 
\begin{itemize}
    \item By categorizing human body parts into three varying levels of DOFs, we explicitly model the bi-directional dependencies among the body parts,  which supervise one another and together yield the physically constrained and plausible 3D pose estimation.
    \item We introduce for each limb joint a pose attribute,
    depicting the offset of the joint from the torso. Estimated together
    with the 2D pose via a multi-task network, the pose attribute
    provides an explicit prior of the joint's location and further facilitates
    the succeeding 3D pose estimation. 
    
\end{itemize}
We test our approach on   
benchmarks including Human3.6M~\cite{ionescu2014human3}
and MPI-INF-3DHP~\cite{mehta2017monocular},
and demonstrate that it consistently
achieves very encouraging results, outperforming the
state of the art.
Furthermore, we show that even
in the absence of 3D annotations and pose-attribute ground truths, 
by adopting an unsupervised domain adaptation approach,
our method can be readily applied to in-the-wild images 
and achieve promising performances.

\section{Related work}
    %According to the way how networks are trained to works on in-the-wild images, existing 3D human pose estimation approaches can be grouped into two categories: one-stage and two-stage approaches.
    
    We briefly review here the two main streams of 3D pose estimation methods,
    the one-stage approaches and the two-stage ones, and then look
    at the methods that explicitly utilize additional image cues. 
    Finally we outline the differences of the propose method 
    with respect to the prior ones.
    
    \paragraph{One-stage approach.} One-stage approaches directly infer 3D human poses from input images.
    Tekin~\etal~\cite{tekin2016structured} trained an auto-encoder to learn a latent pose representation and joint dependencies in a high-dimensional space, then adopted the decoder at the end of the convolutional neural network to infer 3D poses.
    Pavlakos~\etal~\cite{pavlakos2017coarse} proposed a volumetric representation for 3D joints and used a coarse-to-fine  strategy to refine the prediction iteratively.
    Rogez~\etal~\cite{rogez2016mocap,rogez2017lcr} used ConvNets to classify each image in the appropriate pose class. 
    Nie~\etal~\cite{nie2017monocular} proposed to predict the depth on joints from global and local image features.
    
    All the above methods require images with corresponding 3D ground truths.
    Due to the lack of in-the-wild images with 3D annotations, these approaches tend to 
    produce unsatisfactory results on inputs with domain shifts.
    To this end, Zhou~\etal~\cite{zhou2017towards} proposed a weakly-supervised approach to utilize the large-scale in-the-wild 2D pose data.
    Dabral~\etal~\cite{Dabral_2018_ECCV} improved this weakly-supervised setup by using two additional losses to restrict the predicted 3D pose structure. 
    Yang~\etal~\cite{Yang_2018_CVPR} considered the 3D pose estimator as a generator and used an adversarial learning approach to generate indistinguishable 3D poses.  
    Sun~\etal~\cite{Sun_2018_ECCV} used soft argmax to regress 2D/3D poses directly from images.
    Despite the success of this strategy, a main flaw of these methods lies in that they tend to fail when the height of the subject is considerably different from those in the training set, 
    since they fixed the scale of 3D poses to construct 3D poses from 2D poses and depths.
    
    \paragraph{Two-stage approach.} Another widely used strategy is to divide the 3D pose estimation task into two decoupled sub-tasks: 2D pose detection, followed by 
    3D pose inference from 2D poses.
    These methods comprise a 2D pose detector and a subsequent optimization~\cite{zhou2016sparseness,zhou2017sparse,8316924} or regression~\cite{chen20173d,bogo2016keep,moreno20173d,Sun_2017_ICCV,tome2017lifting,martinez2017simple,nie2017monocular,fang2018learning,lee_2018_eccv} step to estimate 3D pose .
    In these methods, the 2D pose and 3D pose estimation stages are separated, 
    making these 3D pose estimators generalize well on outdoor images.
    The most straightforward approach is to represent 3D poses as linear combinations of models learned from training data~\cite{zhou2016sparseness,zhou2017sparse,8316924}.
    This method is based on dictionary learning and has to run an optimization  for each example, making it very time-consuming in both training and evaluation.
    % Also due to the limitation of the model, the accuracy of this method is usually lower compared to deep learning based approaches.
    Specifically, Chen~\etal~\cite{chen20173d} and Yasin~\etal~\cite{yasin2016dual} used a pose library to retrieve the nearest 3D pose given the corresponding 2D pose prediction.
    
    Recently, with the availability of large-scale 3D pose datasets, 
    deep-learning based 2D-to-3D pose regression methods have made significant progress.
    For instance, Moreno-Noguer~\cite{moreno20173d} used an hourglass 
    network to regress the 3D joints distance matrices instead of 3D poses because they found that the distance matrix representation shows a more correlated pattern than Cartesian ones and suffer from smaller ambiguities.
    Notably, Martinez~\etal~\cite{martinez2017simple} achieved state-of-the-art results using a simple multi-layer perceptron with residual blocks~\cite{he2016deep} to regress 3D poses directly from 2D poses.
    Sun~\etal~\cite{Sun_2017_ICCV} re-parameterized the pose presentation to use bones instead of joints and proposed a structure-aware loss. Lee~\etal proposed a long short-term memory~(LSTM) architecture to reconstruct 3D depth from the centroid to edge joints through learning the joint interdependencies.
    However, as 2D-to-3D mapping is an ill-posed problem,  
    methods along this line are prone to ambiguities in the 2D-to-3D regression at the second stage of this pipeline, if no addition image cues are utilized.
  
    \paragraph{Additional image cues.} 
    The idea of pose attributes was firstly explored by Pons-Moll \etal~\cite{pons2014posebits}. In their work, they proposed an extensive set of posebits representing the boolean geometric relationships between body parts, and designed an algorithm to select useful posebits for 3D pose inference.
    Recently, many researchers have investigated 
    approaches that combine 2D pose detection techniques and the power of CNN to extract 
    supplementary information from images to enhance 3D pose estimation.
    Tekin~\etal~\cite{tekin2017learning} proposed a two-stream network with trainable fusion to fuse 2D heat maps and image features to obtain the final 3D pose estimation.
    Pavlakos~\etal~\cite{Pavlakos_2018_CVPR} augmented the LSP and MPII 2D pose datasets with ordinal depth annotations and proposed to learn ordinal depth instead of depth as the compensation information in 2D-to-3D mapping.
    Zhou~\etal~\cite{zhou2017towards} used a CNN to predict 2D joint locations and the corresponding depth, then rescaled the predictions to a pre-defined canonical skeleton.
    All these approaches tried to learn depth information from single images.
    However, an image is a two-dimensional representation itself and does not carry depth information, making it challenging to learn depth from images.
    Also, depth is highly sensitive to camera parameters, such as translation and rotation, making the depth prediction of human joints more difficult.

    \paragraph{Our approach.} 
    By explicitly categorizing body parts 
    into varying levels of DOFs,
    which have been largely overlooked in prior methods,
    the proposed approach treats the 
    higher-DOF parts as dependent components
    of the lower-DOF ones, and conversely, 
    constrains the latter using the former.
    Such bi-directional 3D dependency modeling is further facilitated 
    by a dedicated and newly-introduced pose attribute estimation,  
    which predicts the relative location of a limb joint with respect
    to the torso.

    %Our proposed pose attributes are intrinsic characters of human poses, which are invariant to camera parameters and rigid motion of the subject, making them easier to learn than depth.
    %Given these pose attributes, the size of the solution space of joint in limbs is reduced, and the accuracy of 3D location predictions can be significantly improved.
    %As discussed in the Introduction, previous methods overlooked the differences in difficulty to predict 3D joint locations and treat all the joints equally.
    %We design bi-directional model for learning the joint dependency, which further improves the performance.

\section{Method}
    Different body parts have varying levels of DOFs
    due to the unique  physiological structure of a human body.
    To see this difference, 
    we use the ground truths of the Human3.6M dataset~\cite{ionescu2014human3},
    to compute, for each joint location,
    its standard deviation~(STD), which gives 
    us a coarse description on 
    the motion range of the joint.
    We show the results in Tab.~\ref{tab_joint_std},
    where, as expected, the {distal limb joints}
    including  wrist and ankle have the largest 
    STDs, followed by the {proximal limb joints}
    including elbow and knee.
    The joints on the torso, like spine and hip,
    yield the smallest STDs.
    
    Such DOF differences among body joints lead to the different
    levels of challenges in terms of pose estimation,
    and further result in estimation results of diverse qualities,
    especially those obtained by conventional methods that treat 
    all the parts equally. 
    For example, as shown in Tab.~\ref{tab_error_composition},
    the method of~\cite{martinez2017simple} produces 
    more accurate predictions for joints on the torso and 
    proximal joints on limbs but poorer ones for distal joints.

    \begin{table}[b]
        \small
        \centering
        \begin{tabularx}{0.48\textwidth}{c*{6}{Y}}
            \toprule
            {\bf Joint}   & {\bf Hip}    &{\bf Spine}  &{\bf Thorax}   &{\bf Shoulder}   &{\bf Head}     \\
            {\bf STD (mm)}     &{68.5} &{57.8} &{109}  &{127}  &{140}          \\
            \midrule
            {\bf Joint}   & {\bf Elbow}   & {\bf Knee}   & {\bf Wrist}   & {\bf Ankle}   &
            {\bf Avg.}      \\ 
            {\bf STD (mm)}     &{195}  &{188}  &{240}  &{227}  &{150}          \\
            \bottomrule
        \end{tabularx}
        \caption{The standard deviation of the 3D locations of each joint,
        obtained using the ground-truth annotations of the Human3.6M training set.}
        \label{tab_joint_std}
        \vspace{-4mm}
    \end{table}

    To this end, we categorize the body joints into three levels of DOFs,
    from low to high: \emph{torso}, \emph{proximal limb joints},
    and \emph{distal limb joints}. We then explicitly model the 
    higher-DOF joints as dependent components of the low-DOF ones that are
    easier to estimate, and in turn, enforce the former to impose
    physical constraints on the latter. 
    To aid the learning of this bi-directional dependency,   
    we introduce \emph{pose attribute} to measure the relative
    location of a limb joint with respect to the torso, the body part
    that can be in most cases reliably estimated. 
    Unlike the regression-based depth estimation that is often prone 
    to deviations, the proposed pose attribute estimation is taken to be
    a much less demanding 
    classification problem, where one of the only three labels, \emph{front}, \emph{back},
    and \emph{on-plane},
    is assigned to each limb joint.
    
\begin{figure*}[h]
        \centering
        \includegraphics[width=1\textwidth]{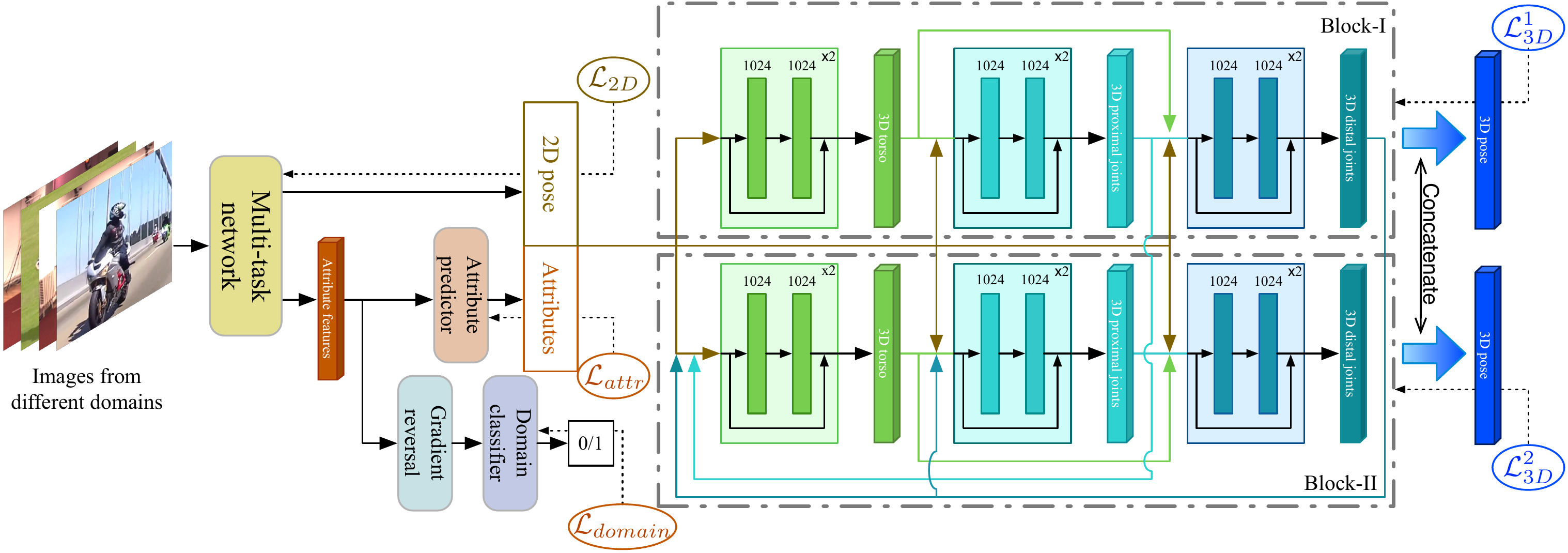}
        \caption{The network architecture of our method.
        It consists of two parts, a multi-task network that learns 2D poses and attributes from images and a progressive 3D pose estimation network.
        The multi-task network is trained on the mixture of MPII and Human3.6M datasets.
        As there are no 3D annotations available in MPII, we adopt an unsupervised domain adaptation method~\cite{ganin2015unsupervised} to help the network learn domain-independent features for attribute prediction, so that the network can predict reasonable attributes for in-the-wild images in the absence of attribute supervision~(see Section~\ref{sec_attr} for more detail).
        The 3D pose network takes as input the estimated 2D poses and pose attributes, and explicitly models the bi-directional dependencies among the three groups of body parts of different DOFs.
        The final 3D pose estimation is the concatenation of the three group predictions.
        }
        \label{fig_overall}
        \vspace{-5mm}
    \end{figure*}

    More specifically, our 3D pose estimation follows a two-step strategy, as depicted in
    Fig.~\ref{fig_overall}. In the first step, we adopt a multi-task 
    network to estimate the 2D pose and the proposed pose attribute, both of which are 
    together fed into another network to model the bi-directional dependency 
    for 3D pose estimation in the second step.
    These two networks are connected via soft argmax layers~\cite{yi2016lift,thewlis2017unsupervised,Luvizon_2018_CVPR,Sun_2018_ECCV},
    so that the network training is end-to-end.  
    In  what follows,  we  provide more details on the two networks.

    \iffalse
    The human skeleton consists of joints with different degrees of freedom.
    The physiological structure of a person determines that the human torso is relatively fixed, while the limbs are flexible.
    % We assume that the joints on the torso are fixed and the other joints have 2 DOF, then the DOF for joints on the torso is 0, and 2 for head, elbows and knees, 4 for wrists and ankles.
    We calculated the standard deviation of per joint locations from the training examples of the Human3.6M dataset (see Tab.~\ref{tab_joint_std}). 
    As expected, the higher the DOF of a joint, the larger the standard deviation of its location in 3D space.
    The significant difference in standard deviations leads to a considerable difference in the difficulty to estimate the the 3D locations of the joints.
    For example, the average error of each joint of a previous method~\cite{martinez2017simple} on Human3.6M dataset is shown in Tab.~\ref{tab_error_composition}.
    We can see that the model tends to generate more accurate predictions for joints on the torso and proximal joints on limbs and less accurate ones for distal joints.
    This phenomenon enlightens us to consider more about the joints on limbs when designing our neural network.
    \fi

    \iffalse
    In order to  predict better 3D limb joints, we propose two techniques, pose attribute learning and progressive regression.
    Pose attribute learning extracts the relative positions between limb joints and the torso plane from image, helping to locate joints in 3D space.
    The other is to explore the mutual joint dependency to refine the 3D pose predictions.
    The joints are divided into three groups according to their DOF, and inferred progressively from low-DOF to high-DOF.

    Correspondingly our 3D human pose estimation network consists of two parts.
    One is the multi-task backbone network, which detects the 2D locations of 16 human joints as well as the pose attributes from input images.
    The other is the head network, which regresses 3D joint locations from low to high DOF with the outputs of the backbone network as input.
    These two networks are connected with soft argmax layers~\cite{yi2016lift,thewlis2017unsupervised,Luvizon_2018_CVPR,Sun_2018_ECCV} instead of the non-differentiable argmax operation so that the network training is end-to-end.  
    In the following sections we will introduce these two networks in detail.
    \fi

    \subsection{Multi-task network}
        The multi-task network,
        as discussed, handles simultaneously  
        2D pose estimation and pose attribute learning.
        In recent years, many network architectures have been proposed
        for 2D pose estimation and have achieved encouraging 
        results~\cite{wei2016convolutional,newell2016stacked,xiao2018simple,Sun_2018_ECCV}.
        Here, we adopt the  state-of-the-art 
        stacked hourglass backbone, as done in many 
        other methods~\cite{martinez2017simple,zhou2017towards,Dabral_2018_ECCV,fang2018learning,Pavlakos_2018_CVPR,Yang_2018_CVPR}, 
        to be our multi-task architecture, following the network design proposed by Zhou~\etal~\cite{zhou2017towards}.
        As the pose attributes are highly related to the locations of joints,
        and a pretrained 2D pose detector can act as a reliable joint feature extractor,
        it is a natural idea to reuse the feature maps in the 2D pose detector to ease pose attribute learning.
        % \xw{, and the 2D detector has learned a good joint feature extractor,
        % it can potentially ease the learning of pose attributes and reduce training time  by reusing the  feature maps of the pretrained hourglass.} \xwc{I don't understand the previous sentence. Please re-word.}
        % The extracted feature maps consists of image features of different abstraction levels, so that the pose attribute learning sub-network can learning from both global and local features.
        
        \subsubsection{2D pose detection}
            Many previous 3D pose estimation methods~\cite{martinez2017simple,fang2018learning,lee_2018_eccv} fine tune a
            pretrained 2D pose detector on the 3D pose dataset like Human3.6M to obtain the 
            pose estimation results.
            Since the images in 3D pose datasets are captured in an indoor lab environment with several subjects,
            the diversity of image backgrounds, clothing, skin color and so on, is very limited compared to the 2D pose datasets.
            As a result, the generalization ability of the model could deteriorate after fine-tuning,
            limiting the application of the pose detector on real-world images.
            % \xwc{Is it a widely accepted fact?}
            
            Here, we train our 2D pose detector from scratch 
            using a mixture of images from both 2D pose and 3D pose datasets.
            In each training batch, half of the examples are 
            randomly sampled from a 2D pose training dataset, and the other from a 3D one.
            Through this strategy, the 2D pose detector could achieves
            high performance on both of the 2D and 3D pose dataset.
            In other words, the 2D pose detector has good generalization capability.
            % \xwc{How to prove this?}
            Moreover, the mixed training strategy also helps to learn pose attributes, as shown in the experiments (see Tab.~\ref{tab_attr_acc}), possibly due to the mixed trained network could learn better human keypoint features. 
            The mixed training strategy is essential in our method.
            On one hand, it helps to train a 2D pose detector with good generalization ability.
            On the other hand, it is also beneficial to the training of the pose attribute learning sub-network by providing better image features.
            % \xwc{We definitely need some experiments without the mixture training.}

            Let us use $M_n$ to denote the ground-truth 2D pose heat map of joint  $n$,
            and use  $\hat{M}_n$ to denote that of the prediction. The loss function 
            for 2D pose detection is taken to be
            \begin{small}
            \begin{equation}
                \mathcal{L}_{2d}=\frac{1}{N}\sum_{n}\text{MSE}(\hat{M}_n-M_n),
            \end{equation}
            \end{small}
            where $N$ is the total number of joints.

        \subsubsection{Pose attributes learning} \label{sec_attr}
            To ease the learning and inference of 3D limb joints, we introduce pose attribute as an additional input for 3D pose estimation.
            The main motivation of introducing such an attribute lies in
            that we aim to encode more visual cues, 
            together with the 2D estimated poses, into the 3D estimation; 
            meanwhile, such visual cues should be estimated reliably.
            To this end, we take pose attribute to be a three-class categorization of 
            the relative location of a limb joint with respect to the torso.
            %  the body part with the least DOF and thus can often be dependably predicted. 
            
            Specifically, we define a torso plane to be the one
            where five body parts lie: left and right shoulder, left and right hip, and the pelvis. In practice, this plane is
            regressed using \emph{Orthogonal Distance Regression},
            where the sum of the Euclidean distances of the five points to the plane is minimized.
            We then compute the Euclidean distances between the obtained 
            torso plane and the joints on the four limbs, including
            left and right shoulder, left and right elbow, 
            left and right knee, left and right ankle, as well as the head.
            Based on the derived distance, a predefined threshold,
            as well as the side of the plane where the joint lies,
            we assign to each join one of the three labels, \emph{front},
            \emph{back}, and \emph{on-plane}.
            The joints with distances smaller than the threshold
            are taken to be on-plane.

            \iffalse         
            %For each joint $i$ in limbs, 
            Assume $d_i$ is the Euclidean distance of joint $i$ to the torso reference plane.
            We define its attribute as:
            \begin{equation*}
                \text{attribute}_i = \left\{
                    \begin{array}{ll}
                        0 & \text{if } d_i<-\tau,\\
                        1 & \text{if } |d_i|\le\tau,\\
                        2 & \text{if } d_i>\tau.\\
                    \end{array}\right.
            \end{equation*}
            where $\tau$ is the interval used to separate ``front", ``on-plane" and ``back".
            \fi

           Let $p_i$ denote the ground-truth probability distribution of the pose attribute on joint $i$, 
           and let $\hat{p}_i$ denote the estimated one. Also, let $\mathcal{J}=$ \{l-Shoulder, l-Elbow, r-Shoulder, r-Elbow, l-Knee, l-Ankle, r-Knee, r-Ankle, head\} denote the set of limb joints.
            Our model predicts all the nine attributes 
            simultaneously with a single network,
            for which a multi-output cross entropy loss is adopted 
            for training:
            \begin{small}
            \begin{equation}
                \mathcal{L}_{attr}=\frac{1}{|\mathcal{J}|}\sum_{i\in\mathcal{J}}\text{CrossEntropy}(\hat{p}_i,p_i),
            \end{equation}
            where $|\mathcal{J}|$ denotes the cardinality of $\mathcal{J}$.
            \end{small}
            
            \iffalse
            Let's denote the ground truth probability distribution of the pose attribute on joint $i$ with $p_i$, the corresponding prediction of probability distribution with $\hat{p}_i$, and the set of indices of limb joints with $\mathcal{J}$, where $\mathcal{J}=$ \{lShoudler, lElbow, rShoulder, rElbow, lKnee, lAnkle, rKnee, rAnkle, head\}. 
            Our model predicts all the nine attributes simultaneously with a single network and a multi-output cross entropy loss is applied:
            \begin{equation}
                \mathcal{L}_{attr}=\frac{1}{|\mathcal{J}|}\sum_{i\in\mathcal{J}}\text{CrossEntropy}(\hat{p}_i,p_i)
            \end{equation}
            \fi

            When training on 2D datasets without 3D annotations, however, 
            we have no attribute supervisions available.
            To deal with this problem, we treat the 2D and 3D 
            training examples as images from different domains,
            and adopt an unsupervised domain adaptation method~\cite{ganin2015unsupervised}
            to help the multi-task network to generate domain-independent features for attribute prediction.
            A classifier is trained to distinguish the domain of the input 
            based on the features to be fed into the attribute predictor,
            while the multi-task network is trained to fool the domain classifier by generating domain-independent features.
            
            Let us denote the ground truth probability distribution of domain with $q$ and the corresponding prediction with $\hat{q}$, the loss function for the domain classifier is taken to be
            \begin{small}
            \begin{equation}
                \mathcal{L}_{domain}=\text{CrossEntropy}(\hat{q},q).
            \end{equation}
            \end{small}

            Specifically, we adopt a gradient reversal layer~\cite{ganin2015unsupervised} to connect the multi-task network and the domain classifier.
            In the forward propagation, the gradient reversal layer acts as an identity function, while in the backward one, it multiplies the gradient by $-\lambda$, where $\lambda>0$.
            As a result, the parameters in the multi-task network are updated in a way to \emph{increase} the loss of the domain classifier,
            which means the CNN tries to learn domain-independent features.
            In the ideal case, the accuracy of the domain classifier is $50\%$, which means attribute features extracted by the multi-task network is not distinguishable at all,
            so it is domain-independent.
            An attribute predictor trained on this domain-independent features is also domain-independent.
            The experiment, as will be demonstrated in Tab.~\ref{tab_attr_acc}, 
            gives strong support to the domain adaptation method above.
            The attribute predictor achieves an accuracy of $84.0\%$ when using this domain adaptation method, and  $82.7\%$ without it.
            The attribute prediction accuracy on the validation set of MPI-INF-3DHP dataset is $70.1\%$ without using any training data from this dataset,
            which also demonstrates the effectiveness of the 
            domain adaptation method.

    \subsection{3D pose estimation network}\label{sec_head}
        
        The 3D pose network takes as input the estimated 2D poses 
        and pose attributes, and explicitly models the bi-directional dependencies 
        among the body parts of different DOFs to produce the final 3D pose estimation.
        By categorizing the joints into three groups, 
        the \emph{torso}, the \emph{proximal limb joints} including the head, the elbows and the knees, and the \emph{distal limb joints} including the wrists and ankles,
        % the torso, the proximal limb joints and the distal limb joints,
        we allow the locations of the higher-DOF 
        groups to be dependent variables 
        of those of the lower-DOF ones, and in turn, 
        constrain the latter using the former.

        Specifically, such bi-directional dependency is achieved via a two-block network architecture, as depicted by the two dotted bounding boxes in Fig.~\ref{fig_overall}.
        Each block models the body parts dependency from one of the two directions.
        Let us denote the 3D joint locations in the three groups as $Y_1$, $Y_2$ and $Y_3$.
        In Block-I, the locations of joints in the lowest-DOF group, $\hat{Y}_{11}$, are inferred from the image evidences learned by the multi-task network, using a basic regression component $G_{11}(\cdot;\theta_{11})$.  
        The prediction results $\hat{Y}_{11}$, due to their low DOF, are usually plausible. 
        The locations of joints $\hat{Y}_{12}$ and $\hat{Y}_{13}$ in the higher-DOF groups,
        are estimated from both the image evidences and their
        dependence upon the predictions of lower-DOF groups.
        For Block-I, we therefore have,
        \begin{small}
        \begin{equation}
            \label{eq_block1}
                \left\{
                \begin{array}{rl}
                    \hat{Y}_{11}&=G_{11}(X;\theta_{11}),\\ 
                    \hat{Y}_{12}&=G_{12}(X,\hat{Y}_{11};\theta_{12}), \\ 
                    \hat{Y}_{13}&=G_{13}(X,\hat{Y}_{12},\hat{Y}_{11};\theta_{13}),
                \end{array}\right.
        \end{equation}
        \end{small}\noindent
        where $X$ denotes the image evidences, $G_{ij}$ denotes the network 
        module that regresses group $j$ in block $i$, and $\theta_{ij}$ 
        represents the learnable parameters in $G_{ij}$.
        
        %At the first block, the one-way dependency from low-DOF to high-DOF is explored.
        %As we have discussed, 
        In Block-II, we enforce the derived high-DOF parts to constrain
        where the low-DOF ones may lie.
        In other words, such dependency is modeled in a reversed direction as
        the one in Block-I. We write,
        \begin{small}
        \begin{equation}
            \label{eq_block2}
                \left\{
                \begin{array}{rl}
                    \hat{Y}_{21}&=G_{21}(X, \hat{Y}_{12}, \hat{Y}_{13};\theta_{21}), \\
                    \hat{Y}_{22}&=G_{22}(X, \hat{Y}_{21}, \hat{Y}_{13};\theta_{22}), \\ 
                    \hat{Y}_{23}&=G_{23}(X, \hat{Y}_{21}, \hat{Y}_{22};\theta_{23}),
                \end{array}\right.
        \end{equation}
        \end{small}\noindent
        where, again, $\hat{Y}_{ij}$, $G_{ij}$, and $\theta_{ij}$
        respectively denote recovered pose locations, network module, and learnable parameters.

        \iffalse
        \begin{figure}[thbp]
            \centering
            \includegraphics[width=0.4\textwidth]{fig/head.pdf}
            \caption{An illustration of the first inference stage.
            }
            \label{fig_progressive}
        \end{figure}
        \fi
        
        The final 3D pose prediction of the  Block-$s$, $\hat{Y}_s$, 
        is the concatenation of all the body parts.
        The loss function is taken to be,
        \begin{small}
        \begin{equation}
            \mathcal{L}_{3D}=\sum_{s\in\{1,2\}}|\hat{Y}_s-Y_s|.
        \end{equation}
        \end{small}\noindent
        Here we choose the $L_1$ loss over $L_2$  as the former shows consistent better performances in our experiments.
        
        % Coincidentally, our network looks similar to the CliqueNet \cite{yang2018convolutional}.
        % The difference is that their network uses convolutional layers to learn a features clique for classification tasks,
        % while ours use fully-conncectly layers to learn a joint clique 3D pose estimation.
        % In Equations~\ref{eq_block1} and~\ref{eq_block2}, $X$ represents the concatenation of detected 2D joint locations and pose attributes.
        % In our network, the earlier predictions are used as the input for later inferences.
        % This is of critical importance in our method, since it helps the network to learn bi-directional dependencies.
        % Without the inter-group connections, even a model with the same amount of parameters performs much worse (see the comparative experiment in Tab.~\ref{tab_error_composition}).

    \begin{table*}[ht]
                \scriptsize
                \begin{tabularx}{\textwidth}{lc*{15}{Y}}
                    \toprule 
                    Protocol \#1                            &Direct.&Discuss&Eating &Greet  &Phone  &Photo  &Pose   &Purch. &Sitting&SittingD&Smoke & Wait  &WalkD  &Walk   &WalkT  &Avg.\\
                    \midrule
                    Tekin~\etal~\cite{tekin2016direct}      &102.4  &147.2  &88.8   &125.3  &118.0  &182.7  &112.4  &129.2  &138.9  &224.9  &118.4  &138.8  &126.3  &55.1   &65.8   &125.0 \\
                    Zhou~\etal~\cite{zhou2016sparseness}   &87.4   &109.3  &87.1   &103.2  &116.2  &143.3  &106.9  &99.8   &124.5  &199.2  &107.4  &118.1  &114.2  &79.4   &97.7   &113.0 \\
                    Du~\etal~\cite{du2016marker}         &85.1   &112.7  &104.9  &122.1  &139.1  &135.9  &105.9  &166.2  &117.5  &226.9  &120.0  &117.7  &137.4  &99.3   &106.5  &126.5 \\
                    Zhou~\etal~\cite{zhou2016deep}         &91.8   &102.4  &96.7   &98.8   &113.4  &125.2  &90.0   &93.8   &132.2  &159.0  &107.0  &94.4   &126.0  &79.0   &99.0   &107.3 \\
                    Chen~\etal~\cite{chen20173d}           &89.9   &97.6   &90.0   &107.9  &107.3  &139.2  &93.6   &136.1  &133.1  &240.1  &106.7  &106.2  &114.1  &87.0   &90.6   &114.2 \\
                    Tome~\etal~\cite{tome2017lifting}      &65.0   &73.5   &76.8   &86.4   &86.3   &110.7  &68.9   &74.8   &110.2  &173.9  &85.0   &85.8   &86.3   &71.4   &73.1   &88.4 \\
                    Rogez~\etal~\cite{rogez2017lcr}         &76.2   &80.2   &75.8   &83.3   &92.2   &105.7  &79.0   &71.7   &105.9  &127.1  &88.0   &83.7   &86.6   &64.9   &84.0   &87.7 \\
                    Pavlakos~\etal~\cite{pavlakos2017coarse}&67.4   &71.9   &66.7   &69.1   &72.0   &77.0   &65.0   &68.3   &83.7   &96.5   &71.7   &65.8   &74.9   &59.1   &63.2   &71.9 \\
                    Nie~\etal~\cite{nie2017monocular}     &90.1   &88.2   &85.7   &95.6   &103.9  &103.0  &92.4   &90.4   &117.9  &136.4  &98.5   &94.4   &90.6   &86.0   &89.5   &97.5 \\
                    Tekin~\etal~\cite{tekin2017learning}    &54.2   &61.4   &60.2   &61.2   &79.4   &78.3   &63.1   &81.6   &70.1   &107.3  &69.3   &70.3   &74.3   &51.8   &74.3   &69.7 \\
                    Zhou~\etal~\cite{zhou2017towards}      &54.8   &60.7   &58.2   &71.4   &62.0   &65.5   &53.8   &55.6   &75.2   &111.6  &64.2   &66.1   &51.4   &63.2   &55.3   &64.9 \\
                    Martinez~\etal~\cite{martinez2017simple}&51.8   &56.2   &58.1   &59.0   &69.5   &78.4   &55.2   &58.1   &74.0   &94.6   &62.3   &59.1   &65.1   &49.5   &52.4   &62.9 \\
                    Sun~\etal~\cite{Sun_2017_ICCV}        &52.8   &54.8   &54.2   &54.3   &61.8   &67.2   &53.1   &53.6   &71.7   &86.7   &61.5   &53.4   &61.6   &47.1   &53.4   &59.1 \\
                    Fang~\etal~\cite{fang2018learning}     &50.1   &54.3   &57.0   &57.1   &66.6   &73.3   &53.4   &55.7   &72.8   &88.6   &60.3   &57.7   &62.7   &47.5   &50.6   &60.4 \\
                    Helge~\etal~\cite{Rhodin_2018_CVPR}     &-      &-      &-      &-      &-      &-      &-      &-      &-      &-      &-      &-      &-      &-      &-      &66.8 \\ 
                    Yang~\etal~\cite{Yang_2018_CVPR}       &51.5   &58.9   &50.4   &57.0   &62.1   &65.4   &49.8   &52.7   &69.2   &85.2   &57.4   &58.4   &\bf43.6   &60.1   &47.7   &58.6 \\
                    Pavlakos~\etal~\cite{Pavlakos_2018_CVPR}&48.5   &54.4   &54.4   &52.0   &59.4   &65.3   &49.9   &52.9   &65.8   &\bf71.1&56.6   &52.9   &60.9   &44.7   &47.8   &56.2 \\
                    Lee~\etal~\cite{lee_2018_eccv}        &\bf43.8&51.7   &48.8   &53.1   &\bf52.2  &74.9   &52.7   &\bf44.6&\bf56.9&74.3   &56.7   &66.4   &68.4   &47.5   &45.6   &55.8 \\
                    Dabral~\etal~\cite{Dabral_2018_ECCV}    &46.9   &53.8   &\bf47.0&52.8   &56.9   &\bf63.6&\bf45.2&48.2   &68.0   &94.0   &55.7   &51.6   &55.4   &40.3   &44.3   &55.5 \\
                    \midrule
                    Ours                                    &44.7   &\bf48.9&\bf47.0&\bf49.0&56.4   &67.7   &48.7   &47.0   &63.0   &78.1   &\bf51.1&\bf50.1&54.5   &\bf40.1&\bf43.0&\bf52.6 \\
                    \toprule
                    Protocol \#2                    & Direct. & Discuss & Eating & Greet & Phone & Photo & Pose & Purch. & Sitting & SittingD & Smoke & Wait & WalkD & Walk & WalkT & Avg\\
                    \midrule
                    Akhter \& Black~\cite{akhter2015pose}   &199.2  &177.6  &161.8  &197.8  &176.2  &186.5  &195.4  &167.3  &160.7  &173.7  &177.8  &181.9  &176.2  &198.6  &192.7  &181.1\\
                    Ramakrishna~\etal~\cite{ramakrishna2012reconstructing}&137.4&149.3&141.6&154.3&157.7&158.9&141.8&158.1  &168.6  &175.6  &160.4  &161.7  &150.0  &174.8  &150.2  &157.3 \\
                    Zhou~\etal~\cite{zhou2017sparse}        &99.7   &95.8   &87.9   &116.8  &108.3  &107.3  &93.5   &95.3   &109.1  &137.5  &106.0  &102.2  &106.5  &110.4  &115.2  &106.7 \\
                    Bogo~\etal~\cite{bogo2016keep}          &62.0   &60.2   &67.8   &76.5   &92.1   &77.0   &73.0   &75.3   &100.3  &137.3  &83.4   &77.3   &86.8   &79.7   &87.7   &82.3 \\
                    Moreno-Noguer~\cite{moreno20173d}       &66.1   &61.7   &84.5   &73.7   &65.2   &67.2   &60.9   &67.3   &103.5  &74.6   &92.6   &69.6   &71.5   &78.0   &73.2   &74.0 \\
                    Pavlakos~\etal~\cite{pavlakos2017coarse}&47.5   &50.5   &48.3   &49.3   &50.7   &55.2   &46.1   &48.0   &61.1   &78.1   &51.1   &48.3   &52.9   &41.5   &46.4   &51.9 \\
                    Martinez~\etal~\cite{martinez2017simple}&39.5   &43.2   &46.4   &47.0   &51.0   &56.0   &41.4   &40.6   &56.5   &59.4   &49.2   &45.0   &49.5   &38.0   &43.1   &47.7 \\
                    Fang~\etal~\cite{fang2018learning}      &38.2   &41.7   &43.7   &44.9   &48.5   &55.3   &40.2   &38.2   &54.5   &64.4   &47.2   &44.3   &47.3   &36.7   &41.7   &45.7 \\
                    Pavlakos~\etal~\cite{Pavlakos_2018_CVPR}&34.7   &39.8   &41.8   &38.6   &42.5   &\bf47.5   &38.0   &36.6   &\bf50.7   &\bf{56.8}&42.6 &39.6   &43.9   &32.1   &36.5   &41.8 \\ 
                    %Yang~\etal~\cite{Yang_2018_CVPR} &\textbf{26.9} &\textbf{30.9} &\textbf{36.3} &39.9 &43.9 &\textbf{47.4} &\textbf{28.8} &\textbf{29.4} &\textbf{36.9} &58.4 &{41.5} &\textbf{30.5} &\textbf{29.5} & 42.5 &\textbf{32.2} &\textbf{37.7} \\
                    Lee~\etal~\cite{lee_2018_eccv}          &38.0   &39.1   &46.3   &44.4   &49.0   &55.1   &40.2   &41.1   &53.2   &68.9   &51.0   &39.1   &56.4   &33.9   &38.5   &46.2 \\
                    Dabral~\etal~\cite{Dabral_2018_ECCV}    &\bf32.8   &\bf36.8   &42.5   &\bf38.5&\bf42.4&49.0   &\bf35.4   &\bf34.3   &53.6   &66.2   &46.5   &\bf34.1   &42.3   &\bf30.0&39.7   &42.2 \\
                    \midrule
                    Ours                                    &33.6   &38.1   &\bf37.6&\bf38.5&43.4   &48.8   &36.0   &35.7   &51.1   &63.1   &\bf41.0&38.6   &\bf40.9&30.3   &\bf34.1&\bf40.7 \\
                    % \midrule
                    \bottomrule
                \end{tabularx}
                 %\vspace{1mm}
                \caption{Detailed results on Human3.6M under Protocol \#1 and \#2.
                All the numbers recorded in the table refer to the mean per joint position errors~(MPJPE) in millimeter.
                The results of all approaches are taken from the original papers.
                Our method outperforms all previous state-of-the-art methods, in terms of the average of the results.
                }
                \label{tab_p1}
                \vspace{-5mm}
            \end{table*}
            
\section{Experiments}\label{sec:experiment}
In this section, we first 
introduce the datasets and protocols we used,
then provide our implementation details,
and next show both the quantitative and qualitative results
as well as the ablation studies. 
{Additional results can be found in our supplementary material.}

    \subsection{Datasets and protocols} \label{sec_protocol}
        We evaluate our method on the following three popular human pose benchmarks.

        \textbf{Human3.6M}~\cite{ionescu2014human3}.
        It contains 3.6 million images and the corresponding 2D pose and 3D pose annotations captured in an indoor environment, featuring $7$ subjects performing $15$ everyday activities like ``Eating" and ``Walking".
        We follow the standard protocol on Human3.6M to use S1, S5, S6, S7 and S8 for training and S9 and S11 for evaluation.
        The evaluation metric is the mean per joint position error~(MPJPE) in millimeter between the ground-truth and the prediction across all cameras and joints after aligning the depth of the root joints.
        We refer to this as Protocol \#1.
        In some works, the predictions are further aligned with the ground-truth via a rigid transformation.
        We refer to this as Protocol \#2.
        Following~\cite{zhou2016sparseness,pavlakos2017coarse,zhou2017towards,Pavlakos_2018_CVPR}, we down sampled the original videos from 50fps to 10fps to remove redundancy.
        We employed all camera views and trained a single model for all activities.

        \textbf{MPII}~\cite{andriluka20142d}. It is the most widely used benchmark for 2D human pose estimation.
        It contains 25K in-the-wild images collected from YouTube videos covering a wide range of activities.
        It provides 2D annotations but no 3D ground truth.
        As a result, direct image-to-3D training is not a practical option with this dataset.
        We adopt this dataset for the training and testing of the multi-task network and for the qualitative evaluation of our 3D pose estimation method.

        \textbf{MPI-INF-3DHP}~\cite{mehta2017monocular}. It is a recently proposed 3D pose dataset constructed by the Mocap system with both constrained indoor scenes and complex outdoor scenes.
        We only use the test split of this dataset, which contains $2929$ frames from six subjects performing seven actions, to evaluate the generalization ability quantitatively and qualitatively.

        \subsection{Implementation details}
        Our method is implemented using PyTorch~\cite{paszke2017automatic}.
        The training procedure of our network consists of three steps: training the multi-task network, training progressive regression network, and connecting them and fine-tuning.
        For the first step, the multi-task network is trained for 60 epochs. 
        The learning rate is set to $5\times10^{-4}$ and batch size is 12. 
        For the second step, the 3D pose regression network is trained on predicted 2D key point positions and ground-truth pose attributes for 60 epochs.
        The learning rate is set to $2.5\times10^{-4}$ and batch size is 64. 
        For the final step, the multi-task network and the 3D pose network are connected with soft argmax layers and fine-tuned for 40 epochs.
        The learning rate is set to $1.0\times10^{-4}$ and batch size is 64.
        The first and third steps  are trained on the mixture of MPII and Human3.6M datasets.
        The training examples are randomly sampled from the two datasets with equal probability.
        Augmentation of random scale ($1\pm0.2$) and random color jitter ($1\pm0.2$) are used for both datasets.
        For the MPII dataset, random rotation ($\pm30^{\circ}$) and random horizontal flipping are also used.
        The RMSprop optimizer is used for all the training steps.
        The whole training procedure takes about 2 days on two Tesla V100 GPUs with 16G memory for each.

    \subsection{Quantitative results}
    In what follows, we show our quantitative results on 2D pose estimation, 
    on attribute prediction, on 3D pose estimation, 
    qualitative results, and ablation studies. 
    
        \subsubsection{2D pose estimation results on MPII}
            The accuracy of 2D pose detection is known 
            to be crucial for 3D estimation~\cite{martinez2017simple}.
            Although our 2D detector is trained on the mixture of MPII and Human3.6m dataset, the PCKh@0.5 score of our model on the MPII validation split is very close to previous works~\cite{Yang_2018_CVPR} (see Tab.~\ref{tab_2D_acc}).
            This indicates that the performance improvement does not rely on a extremely well-trained 2D detector.
            \begin{table}[t]
                \small
                \begin{tabularx}{0.48\textwidth}{l*{8}{Y}}
                    \toprule
                    Model                   & Head  &Sho.   &Elb.   &Wri.   &Hip    &Knee   &Ank.   &Avg.   \\ 
                    \midrule
                    HG                      &{96.3} &{95.0} &{89.0} &{84.5} &{87.1} &{82.5} &{78.3} &{87.6} \\
                    \cite{Yang_2018_CVPR}   &{96.1} &{95.6} &{89.9} &{84.6} &{87.9} &{84.3} &{81.2} &{88.6} \\
                    Ours                    &{94.7} &{94.0} &{90.8} &{88.7} &{84.9} &{83.0} &{83.4} &{88.5} \\
                    \bottomrule
                \end{tabularx}
                 %\vspace{1mm}
                \caption{PCKh@0.5 score on the MPII validation set.
                Joints are grouped by bilateral symmetry (ankles, wrists, etc).
                HG represents the pretrained hourglass model~\cite{newell2016stacked}.
                The 2D pose detection performance of our multi-task network is very close to that of~\cite{Yang_2018_CVPR},
                while our 3D results are much better (see Tab.~\ref{tab_p1} and Tab.~\ref{tab_MPI_INF_3DHP}).
                }
                \label{tab_2D_acc}
                \vspace{-5mm}
            \end{table}
        \subsubsection{Performance of attribute prediction}
            In this section, we conduct experiments to find out the best training strategy to learn the attributes.
            There are three candidate training strategies, training on only Human3.6M, training on the mixture of MPII and Human3.6M,
            and training on the mixture with domain adaptation~(DA).
            From Tab.~\ref{tab_attr_acc} we can see that the mix-training strategy can significantly boost the attribute prediction accuracy. 
            With the help of the domain adaptation, the attribute prediction is further improved. 
            
            It is worth noting that our attribute predictor also works well on the MPI-INF-3DHP dataset without using any examples from this dataset for training, which shows that our multi-task network successfully learns to transfer between domains.
            \begin{table}[t]
                \small
                \begin{tabularx}{0.48\textwidth}{llc*{5}{Y}}
                    \toprule
                    Dataset     &Method &Head   &Elb.   &Wri.   &Knee   &Ank.   &Avg.\\ 
                    \midrule
                    {}          &h36m   &75.7   &77.2   &80.6   &82.0   &77.9   &78.6\\
                    {H36M}      &mix    &79.2   &80.9   &87.5   &84.0   &82.1   &82.7\\
                    {}          &mix+DA &79.4   &82.6   &88.4   &85.9   &83.6   &\bf84.0\\
                    \midrule
                    {}          &h36m   &{47.5} &{48.4} &{58.7} &{59.6} &{41.4} &{51.1}\\
                    {MPI3D}     &mix    &{74.6} &{67.1} &{72.5} &{69.2} &{55.3} &{67.7}\\
                    {}          &mix+DA &{73.1} &{65.0} &{71.1} &{79.7} &{61.8} &\bf70.1\\
                    \bottomrule
                \end{tabularx}
                 %\vspace{1mm}
                \caption{The accuracy of attribute prediction on the Human3.6M~(H36M)  and the MPI-INF-3DHP~(MPI3D) dataset. 
                \emph{H36m} stands for training  using only  Human3.6M, \emph{mix} stands for using the mixture of Human3.6M and MPII, and \emph{DA} stands for using the domain adaptation method discussed in Section~\ref{sec_attr}. No training data from MPI-INF-3DHP have been used for training.
                }
                \label{tab_attr_acc}
            \end{table}
            
               \begin{figure*}[t]
                \centering
                \includegraphics[width=1.0\textwidth]{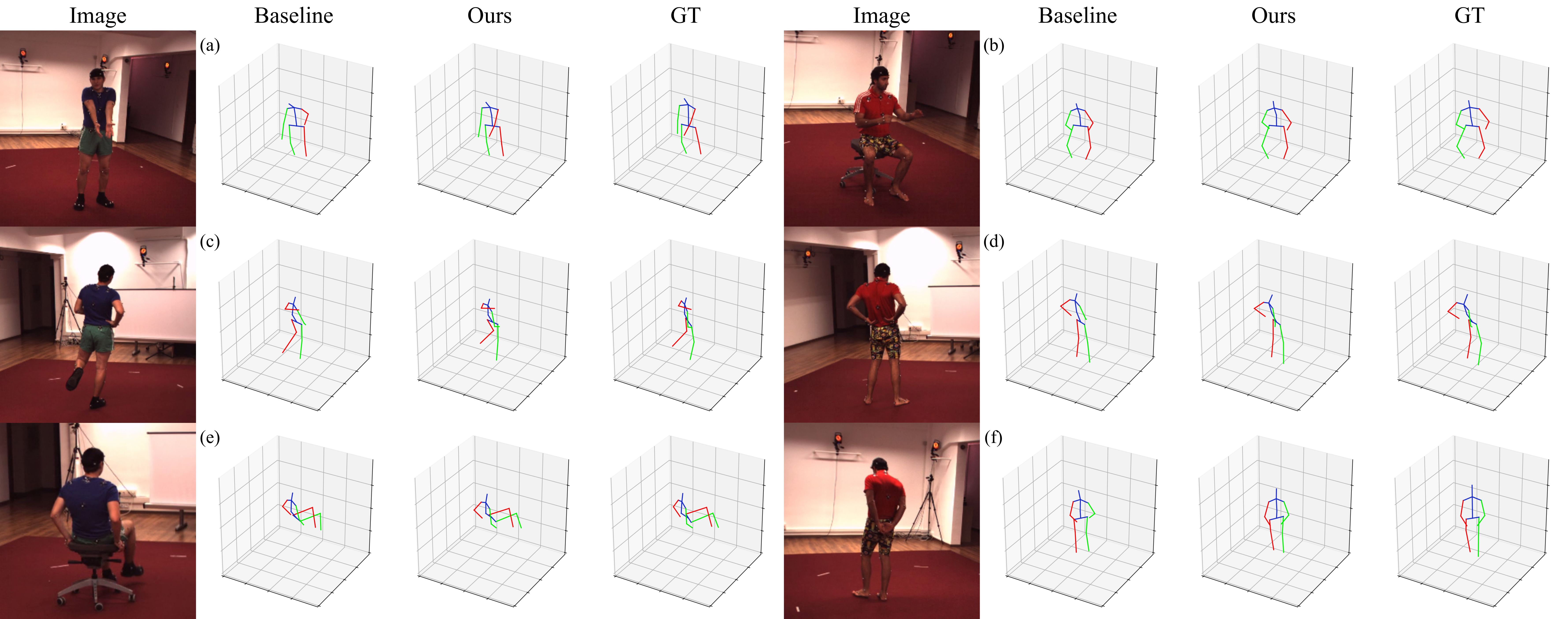}
                \caption{Qualitative results on the Human3.6M dataset.
                Our predictions of limbs are significantly better than those of the baseline,~defined in Section~\ref{sec_ablation}.
                }
                \label{fig_quali_h36m}
                % It shows that our method works well on in-the-wild images, even when the 2D pose is severely ambiguous (see the examples at first two rows).
                % The third row is some examples that have no significant ambiguity. Both the baseline and the proposed method work well on these examples.
                \vspace{-3mm}
            \end{figure*}
            
        \subsubsection{3D pose estimation results on Human3.6M}
            We evaluate our method using the two most popular protocols (see Section~\ref{sec_protocol}) on Human3.6M.
            The detailed results of our method and previous state-of-the-art methods are listed in Tab.~\ref{tab_p1}.
            Our method outperforms previous methods, in terms of 
            the average result on all the actions.
            
        \subsubsection{3D pose estimation results on MPI-INF-3DHP}
            We evaluate our method on another unseen 3D human pose dataset, MPI-INF-3DHP~\cite{mehta2017monocular}, to test the cross-domain generalization ability. We follow~\cite{mehta2017monocular,mehta2017vnect,zhou2017towards,Yang_2018_CVPR,Pavlakos_2018_CVPR} to use 3DPCK and AUC as the evaluation metrics.
            Comparisons with previous methods are shown in Tab.~\ref{tab_MPI_INF_3DHP}.
            Our method outperforms the prior ones on this unseen dataset, demonstrating
            the robustness of our method on domain shift.
            \begin{table}[t]
                \small
                \begin{tabularx}{0.48\textwidth}{l*{5}{Y}}
                    \toprule
                    {} & \cite{mehta2017monocular} & \cite{zhou2017towards} & \cite{Pavlakos_2018_CVPR} & \cite{Yang_2018_CVPR} & ours\\ 
                    \midrule
                    3DPCK & {64.7} & {69.2} & \bf{71.9} & {69.0} & \bf{71.9} \\
                    AUC & {31.7} & {32.5} & {35.3} & {32.0} & \bf{35.8} \\
                    \bottomrule
                \end{tabularx}
                 %\vspace{1mm}
                \caption{3DPCK and AUC on the MPI-INF-3DHP dataset.
                The results for all approaches are taken from the original papers.
                No training data from this dataset have been used for training.}
                \label{tab_MPI_INF_3DHP}
                \vspace{-3mm}
            \end{table}

    \subsection{Qualitative results}
\begin{figure}[htbp]
            \centering
            \includegraphics[width=0.45\textwidth]{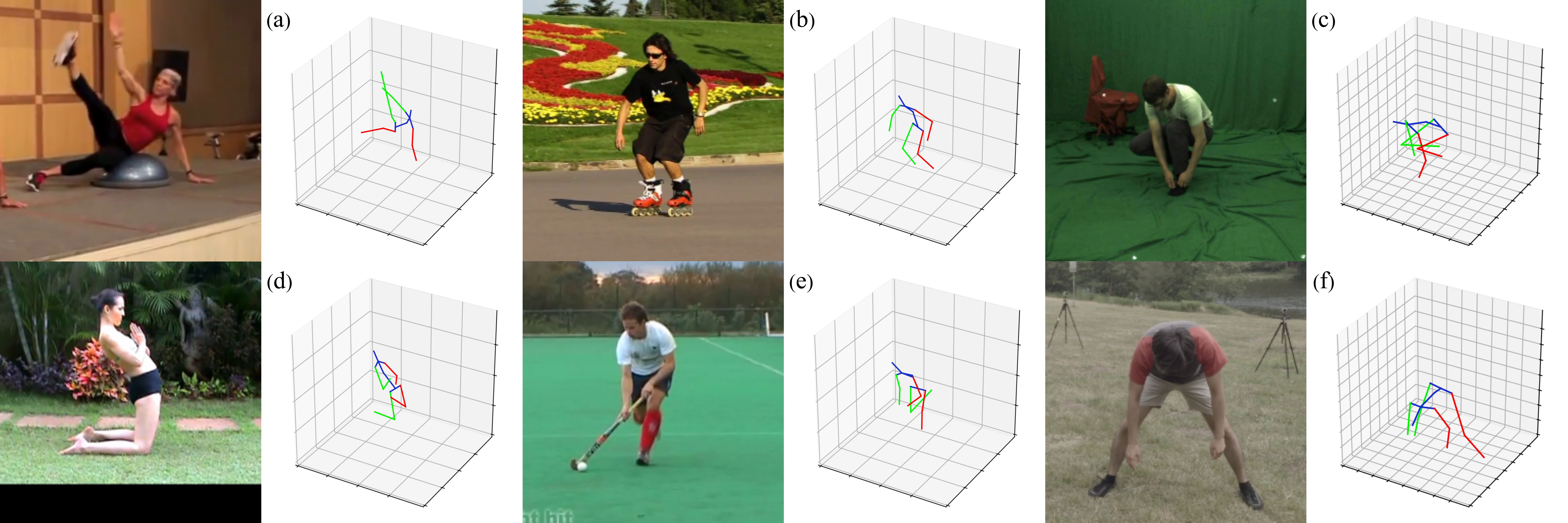}
            \caption{Qualitative results on datasets with domain shift. The first two columns are from MPII and the last is  from MPI-INF-3DHP.
            }
            \label{fig_mpii}
                % It shows that our method works well on in-the-wild images, even when the 2D pose is severely ambiguous (see the examples at first two rows).
                % The third row is some examples that have no significant ambiguity. Both the baseline and the proposed method work well on these examples.
            \vspace{-4mm}
        \end{figure}

        % \subsubsection{3D pose predictions on indoor examples from Human3.6m}
        % compare baseline, progressive, progressive + attribute
        % \subsubsection{3D pose predictions on in-the-wild examples from MPII}
        In Fig.~\ref{fig_quali_h36m} we show 
        visualizations of several 3D pose predictions of our method on Human3.6M.
        As we may observe, our results are visually very close to the ground truths
        and considerably better than the baseline approach to be discussed in 
        Section~\ref{sec_ablation}.
        Besides, in Fig.~\ref{fig_mpii} we give the qualitative results on images in other scenes, including those from MPII and MPI-INF-3DHP, 
        to show the robustness of our method to domain shift.
        % A richer collection of success and failure examples is included in the supplementary material.
            
    \subsection{Ablation studies} \label{sec_ablation}
        To analyze the effectiveness of each component, we conduct ablation study on Human3.6M under Protocol \#1.
        The mean per joint error is reported in Tab.~\ref{tab_error_composition}.
        The notations are defined as follows:
        \begin{itemize}
            \item \textbf{Baseline} 
            refers to the approach that adopts the same network architecture as ours, 
            but without modeling  bi-directional dependencies 
            among body parts and without pose attribute as input.
           % We thus have            as defined in 
            %Eq.~\ref{eq_baseline}, 
            % Eqs.~(\ref{eq_block1})~and~(\ref{eq_block2}), \xwc{Is this what you mean?} 
            %to model the bi-directional dependency. 
            In other words, we model only,
            \begin{small}
            \begin{equation}
                 Y_n = G_n(X;\theta_n) \quad n\in\{1,2,3\}.
                 \label{eq_baseline}
            \end{equation}
            \end{small}
            \item \textbf{Progressive} refers to the  bi-directional approach, as introduced
            in Section \ref{sec_head}, but without  pose attribute as input.
            \item \textbf{Attr} refers to using the predicted pose attributes to estimate the 3D pose.
        \end{itemize}
        We also re-implemented the method in~\cite{martinez2017simple} for comparison.
        Our implementation in fact yields results slightly better than
        those reported in the original paper.% \xwc{is this what you mean?}

        \begin{table}[ht]
            \small
            \begin{tabularx}{0.48\textwidth}{l*{6}{Y}}
                \toprule
                Joint                       &Hip    &Spine  &Thorax &Shoulder   &Head\\
                \midrule
                \cite{martinez2017simple}   &20.7   &37.6   &42.5   &56.5       &65.3\\
                \midrule
                Baseline                    &20.9   &38.3   &43.0   &56.4       &65.0\\
                Progressive                 &21.4   &37.9   &42.8   &56.0       &63.6\\
                
                Progressive + Attr          &\bf20.4&\bf36.8&\bf40.6&\bf52.2    &\bf58.4\\
                
                \toprule
                Joint                       &Elbow  &Knee   &Wrist  &Ankle  &Avg.\\
                \midrule
                \cite{martinez2017simple}   &81.6   &58.7   &100.3  &84.8   &59.4\\
                \midrule
                Baseline                    &80.6   &56.4   &98.5   &81.9   &58.3\\
                Progressive                 &78.4   &55.7   &94.2   &80.1   &56.9\\
                Progressive + Attr          &\bf71.3&\bf51.0&\bf87.6&\bf74.7&\bf52.6\\
                \bottomrule
            \end{tabularx}
             %\vspace{1mm}
            \caption{The prediction errors of \cite{martinez2017simple}
            and our model by turning some modules off.}
            \label{tab_error_composition}
            \vspace{-4mm}
        \end{table}
        
        \iffalse
        \begin{table}[ht]
            % \small
            \scriptsize
            \begin{tabularx}{0.48\textwidth}{l*{9}{Y}}
                \toprule
                Joint       &Spi.   &Tho.   &Head   &Sho.   &Elb.   &Wri.   &Hip    &Knee   &Ank.\\ 
                \cite{martinez2017simple} &37.6 &42.5   &65.3   &56.5 &81.6 &100.3 &20.7 &58.7 &84.8 \\
                \midrule
                Baseline    &{38.3} &{43.0} &{65.0} &{56.4} &{80.6} &{98.5} &{20.9} &{56.4} &{81.9}\\
                Prog        &{37.9} &{42.8} &{63.6} &{56.0} &{78.4} &{94.2} &{21.4} &{55.7} &{80.1}\\
                Prog + attr &{36.8} &{40.6} &{58.4} &{52.2} &{71.3} &{87.6} &{20.4} &{51.0} &{74.7}\\
                \bottomrule
            \end{tabularx}
            \caption{The errors of each joint}
            \label{tab_error_composition}
        \end{table}
        \fi
        Although the number of parameters of the baseline model and the progressive model are almost the same, the performance of the later is significantly better.
        From this comparative experiment we can see that  the bi-directional model is indeed effective in 3D pose estimation.
        The proposed attributes further boosts the performance by a large margin, especially on the the joints where a pose attribute is defined, proving the effectiveness the proposed pose attributes.

\section{Conclusion}

    We propose in this paper a two-step 3D pose estimation approach that explicitly models the bi-directional dependencies among body parts of different DOFs. In the first step, we adopt a multi-task network that jointly estimates the 2D poses and the pose attributes for each limb joint,
    a three-class categorization that depicts the relative location between
    a joint and the torso plane. 
    The pose attribute, unlike the more challenging regression-based depth estimation, 
    provides a dependable yet informative prior of the joint locations.
    The  predictions of 2D poses and attributes are then fed to the 3D pose estimation network, where higher-DOF parts are explicitly modeled as
    dependent variables of lower-DOF parts and meanwhile constrain the locations
    of the lower-DOF ones. 
    In this way, body parts of different DOFs supervise and benefit one another,
    together yielding the encouraging results that outperform the state of the art on standard benchmarks. 
    %\xwc{Any other interesting future work?}

   % This paper presents an alternative perspective that utilizes pose attributes to help locate the joints on limbs in 2D-to-3D pose regression.
    %These attributes are reliable compensation for missing depth information, and are easy to learn from images.
    %We demonstrated the capability of the proposed pose attributes to reduce the ambiguities in 2D-to-3D pose regression, by incorporating them in an 3D pose estimation component. 
    %The proposed bi-directional model for 3D pose regression has been proven to be effective to explore the mutual dependency among joints.
    
    %Our accuracy in 3D pose estimations suggests that it is a promising way to explore mutual joint dependency to learn 3D poses.
    %The pose attribute is a simple attempt to explore more image evidences other than 2D poses to predict better 3D poses.
    %Currently it only looks at the limbs and heads.
    %One possible future work is to learn better image cues for example full-body attributes. 
    %On the other hand, our architecture is perhaps the simplest architecture one may think of to learn the mutual joint dependency.
    %We believe that a further exploration of the network architectures will result in improved performance.
    %These are all interesting areas of future work.

{\small
\bibliographystyle{ieee}
\bibliography{ref}
}

\end{document}